\documentclass[conference, 10pt, comsoc]{IEEEtran}
\IEEEoverridecommandlockouts
\usepackage{cite}
\usepackage{amsmath,amssymb,amsfonts}
\usepackage{algorithmic}
\usepackage{graphicx}
\usepackage{svg}
\usepackage{textcomp}
\usepackage{xcolor}
\usepackage{booktabs}
\def\BibTeX{{\rm B\kern-.05em{\sc i\kern-.025em b}\kern-.08em
    T\kern-.1667em\lower.7ex\hbox{E}\kern-.125emX}}
\begin{document}

\title{RAG-based User Profiling for Precision Planning in Mixed-precision Over-the-Air Federated Learning\\

\thanks{The authors are with the Faculty of Engineering and Applied Sciences, Cranfield University, United Kingdom. The work is supported by EPSRC CHEDDAR: Communications Hub for Empowering Distributed clouD computing Applications and Research (EP/X040518/1) (EP/Y037421/1).}
}

\author{
\IEEEauthorblockN{Jinsheng Yuan}
\IEEEauthorblockA{jinsheng.yuan@cranfield.ac.uk}
\and
\IEEEauthorblockN{Yun Tang}
\IEEEauthorblockA{yun.tang@cranfield.ac.uk}
\and
\IEEEauthorblockN{Weisi Guo}
\IEEEauthorblockA{weisi.guo@cranfield.ac.uk}
}

\maketitle

\begin{abstract}
Mixed-precision computing, a widely applied technique in AI, offers a larger trade-off space  between accuracy and efficiency. The recent purposed Mixed-Precision Over-the-Air Federated Learning (MP-OTA-FL) enables clients to operate at appropriate precision levels based on their heterogeneous hardware, taking advantages of the larger trade-off space while covering the quantization overheads in the mixed-precision modulation scheme for the OTA aggregation process. A key to further exploring the potential of the MP-OTA-FL framework is the optimization of client precision levels. The choice of precision level hinges on multifaceted factors including hardware capability, potential client contribution, and user satisfaction, among which factors can be difficult to define or quantify.

In this paper, we propose a RAG-based User Profiling for precision planning framework that integrates retrieval-augmented LLMs and dynamic client profiling to optimize satisfaction and contributions. This includes a hybrid interface for gathering device/user insights and an RAG database storing historical quantization decisions with feedback. Experiments show that our method boosts satisfaction, energy savings, and global model accuracy in MP-OTA-FL systems.

\end{abstract}


\section{Introduction}

Over-the-Air Federated Learning (OTA-FL) \cite{Yang2020OTAFL} represents a novel paradigm in FL dedicated to wireless networks, which leverages the inherent randomness of physical-layer channel states and electromagnetic superposition for aggregating model updates. The same property is utilized to accommodate model parameters of multiple computational precision in Mixed-Precision OTA-FL \cite{yuan2024MPOTAFL}. The approach enables clients with heterogeneous hardware to participate in FL with better trade-offs between performance and energy efficiency than homogeneous precision FL systems while covering the quantization overheads in mixed-precision OTA aggregation.

\begin{figure}[htbp]
    \centering
    \includegraphics[width=\columnwidth]{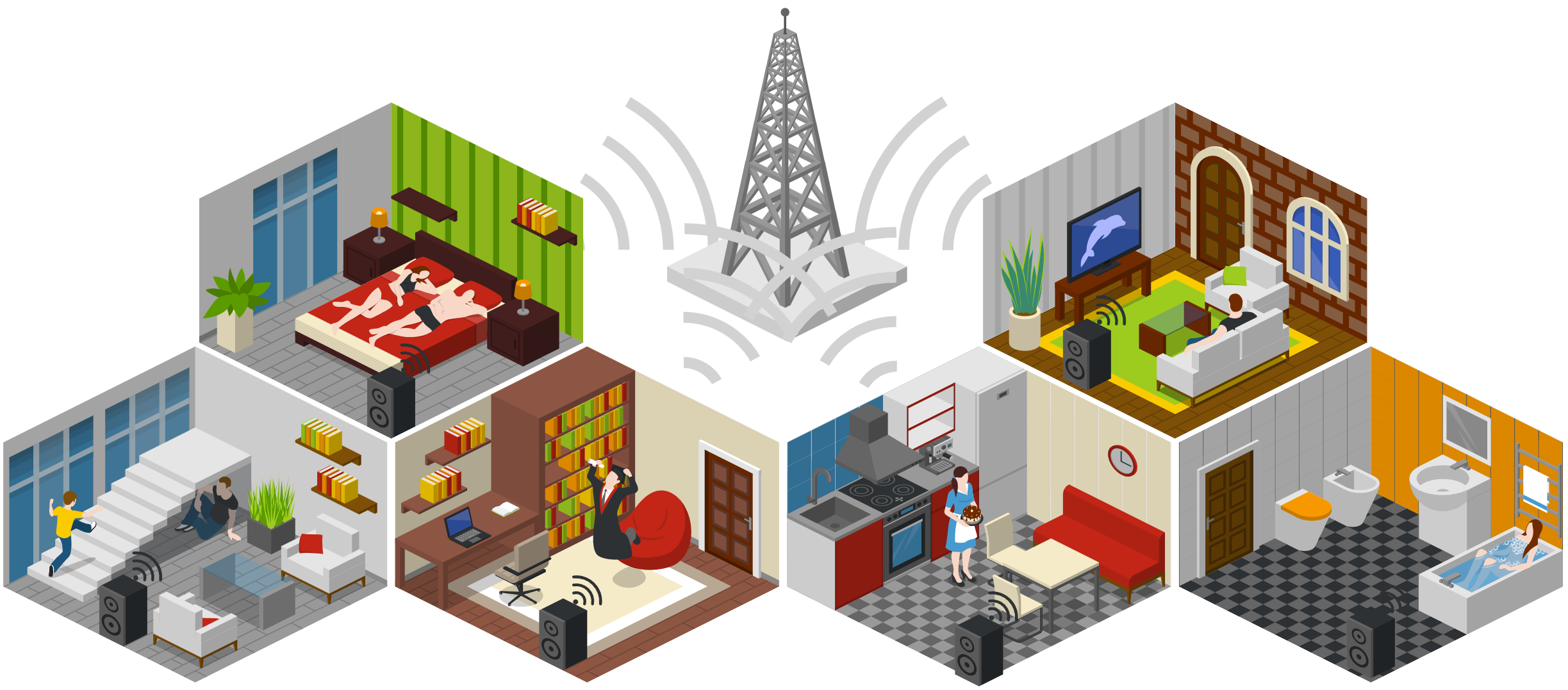}
    \caption{User satisfaction and client contribution potentials in federated learning vary with contextual factors such as usage patterns and operational environment.}
    \label{fig:different_contextual_factors}
\end{figure}

One of the keys to enhancing such mixed-precision OTA-FL systems is to select the optimal quantization level for each client, which hinges on multifaceted factors within two main themes, user satisfaction, and client contribution. For user satisfaction, since available quantization levels depend on hardware specification, and the quantization to different levels directly translates to the corresponding performance metrics (e.g., accuracy, delay, energy efficiency), the most satisfying precision level can vary largely due to different usage patterns and performance sensitivity among users, even for those with identical hardware. As for client contribution, which depends on the quality, quantity, and distribution of client data, is infeasible to quantify directly due to the opacity of the client dataset. While there exist various proxy approaches for contribution estimation~\cite{Jia2021saclability,jia2019SV}, such estimations are often limited by the impractical demand for additional computation or the invalidation of assumptions in real-world scenarios.

To plan the optimal quantization level for each client, with main considerations of user satisfaction and client contribution, it's essential to collect and assess both the intrinsic technical factors such as hardware specifications (e.g., power states, compute capacity), model performance under quantization, and the extrinsic contextual factors such as device operational environments, user-specific usage patterns and preferences (as shown in Fig.\ref{fig:different_contextual_factors}). However, collecting and assessing such factors poses the following challenges:

\begin{itemize}

    \item \textbf{Challenge 1: Difficulty in comprehensive enumeration of extrinsic factors.} While intrinsic factors such as hardware specifications and quantized model performance can be systematically captured, extrinsic factors, including operational environments (e.g., ambient noise levels) and usage patterns (e.g., engagement frequency, input data classes) and user satisfaction are inherently highly dynamic and multifaceted, making exhaustive enumeration challenging and necessitating nuanced approaches.

    \item \textbf{Challenge 2: Complexity in quantifying individual client contribution.} Direct assessment of a client's potential contribution to global model accuracy relies on client data characteristics including quantity, quality, and distribution, which are inherently inaccessible in FL due to user privacy. Existing proxy methods often involve client exclusive testing, and hence limited by the resulted computational overhead, highlighting the need for adaptive, context-aware estimation strategies.
    
\end{itemize}

The rapid-advancing Large Language Models (LLMs) offer a promising approach to address these challenges. In this paper, we propose a \textbf{RAG-based User Profiling for Precision Planning Framework}, integrating retrieval-augmented LLMs with dynamic client profiling to enhance user satisfaction. Specifically, to address Challenge 1, we develop a hybrid conversational interface combining available hardware information for capturing resource constraints with an interactive LLM-driven conversational agent to identify latent user needs and operational contexts. To address Challenge 2, we establish a knowledge database using Retrieval-Augmented Generation (RAG), which maintains historical quantization planning records alongside associated user feedback, thereby creating semantic mappings between contextual factors and user factors including satisfaction and contribution potentials to global model. Following the comprehensive collection of these factors, we compute reward-penalty metrics for each client's precision levels to optimize precision selection for subsequent learning rounds. User feedback gathered during this process is continuously integrated into the knowledge database, facilitating continuous refinement in precision planning.
The contributions of this paper are as follows:

\begin{enumerate}
    \item A RAG knowledge database that semantically links historical quantization decisions and user feedback, enabling data-driven estimation of the effect of contextual factors on user satisfaction and global model accuracy.
    
    \item A dynamic client profiling mechanism that leverages an LLM agent-powered chat interface to extract user preference and contextual factors.
    
    \item An experimental demonstration of the effectiveness of the framework through a mixed-precision FL voice assistant system in terms of user satisfaction, energy consumption and accuracy.
    
    \item A open-sourced framework implementation \footnote{https://github.com/ntutangyun/user\_in\_the\_loop\_quantization\_planning} for the community to use, adapt and contribute.
\end{enumerate}

The rest of the paper is organized as follows. Section II reviews related works. Section III presents the proposed framework. Section IV describes the experimental setup and results. Section V concludes the paper.

\section{Related Works}

\subsection{Mixed-Precision Federated Learning}

Federated learning, since its introduction by McMahan et al. \cite{McMahan2016FL}, has been widely applied in privacy-sensitive distributed computing scenarios such as healthcare, finance, and IoT. Mixed-precision computation has been widely employed in deep learning, improving efficiency in both training and inference~\cite{micikevicius2017mixed}. The insight behind such design is that different types of layers in neural networks have different sensitivity to computation precision. Generalizing from this, mixed-precision OTA FL~\cite{yuan2024MPOTAFL}, with quantization overheads covered by OTA aggregation, offers a larger trade-off area between precision and performance, especially for those clients operating at the lowest precision levels due to most limited resources.

\subsection{RAG-LLMs}

RAG-LLMs, introduced by Lewis et al. \cite{lewis2020retrieval}, are a class of large language model that generate responses based on retrieved relevant information from external knowledge sources. RAG-LLMs have been deployed and achieved impressive performance in various NLP tasks such as user profiling for recommendation systems~\cite{deldjoo2024genrecsys}.

\section{Methodology}

\subsection{Framework Overview}
\label{subsec:framework_overview}

\begin{figure*}[ht]
    \centering
    \includegraphics[width=0.9\textwidth]{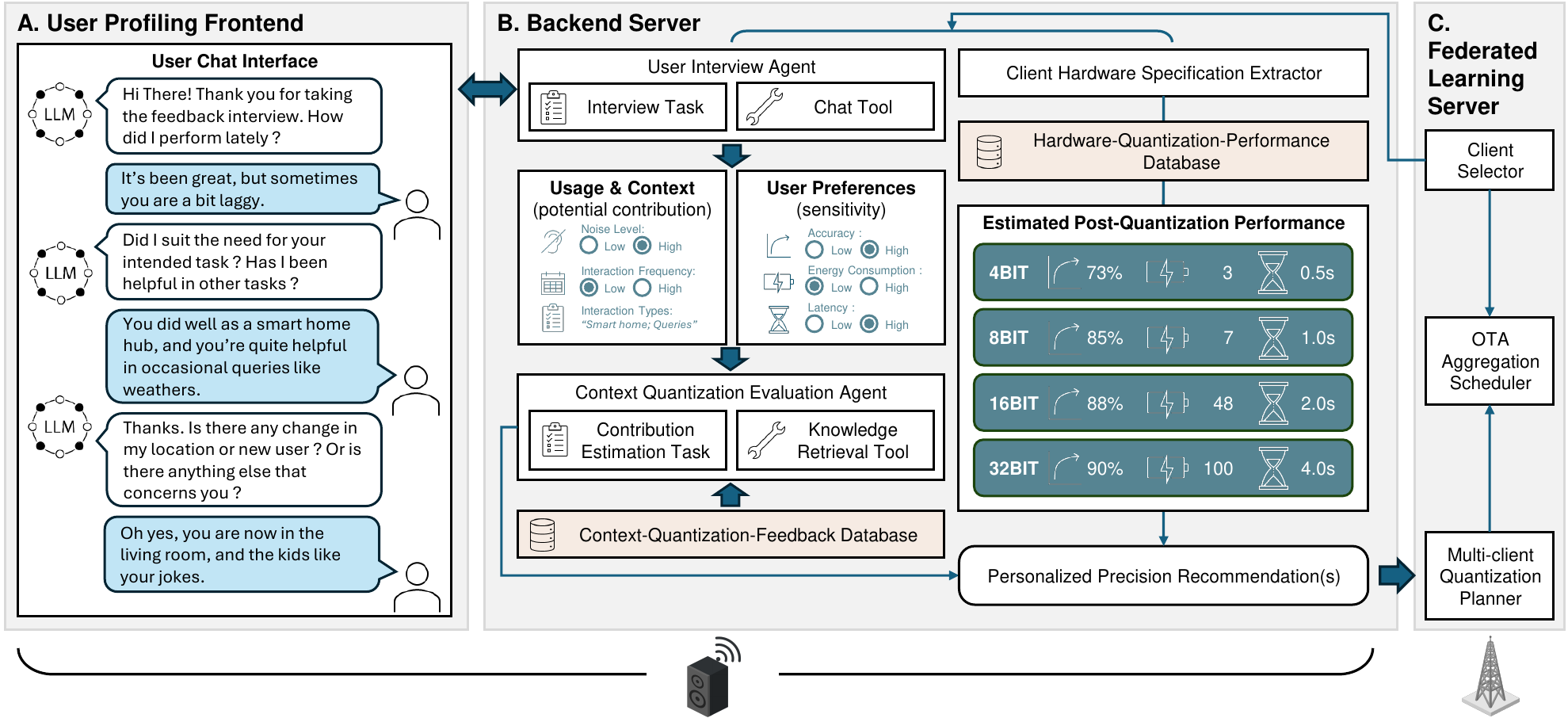}
    \caption{User-in-the-loop Quantization Planning Framework Overview. The aim is to collect the user's feedback on the $T$ round and select the optimal quantization level for the $T+1$ round for the federated learning process.}
    \label{fig:framework-overview}
\end{figure*}

The proposed precision planning framework adopts a full-stack architecture as illustrated in Fig.~\ref{fig:framework-overview}. The framework comprises two core components: a user-profiling frontend featuring a chat interface and a backend server hosting an LLM-powered agent. 

\textbf{User Profiling Frontend} A LLM agent-driven conversational interface for contextual factor discovery and satisfaction feedback collection. The conversation is tasked by the backend, and the primary tasks are as follows:
\begin{itemize}
    \item At new device initialization, the user is prompted to provide perspective usage patterns and setup contexts, e.g., device location, intended usage scenarios, and user preferences.
    \item At the pre-aggregation stage, the user is queried for feedback on the performance of the operation, as well as potential context change, since the last feedback collection.
    \item In the case of changed hardware specifications, the user is prompted to update contextual factors and preferences.
\end{itemize}

\textbf{Quantization Optimization Backend:} A knowledge-enhanced processing stack containing:
\begin{itemize}
    \item A RAG knowledge database (Context-Quantization-Feedback Database) that archives precision decision history, including usage patterns, operational contexts, and corresponding user feedback.
    \item A knowledge database (Hardware-Quantization-Performance Database) that archives model performance (e.g., accuracy) with the associated hardware and precision level.
    \item LLM interview agent that interviews user preferences (e.g., user sensitivity for accuracy, response time and energy consumption) and usage patterns (e.g., noise level, usage frequency and type of interactions).
    \item Hardware specification extractor that collects device hardware information based on availability and user privacy settings. The parsed hardware specs are then used to query the knowledge database to estimate quantization-performance trade-offs on similar user hardware.
    \item Context quantization evaluation agent that estimates the potential client contribution and user satisfaction at available precision levels based on interpreted contextual factors and retrieved hardware capabilities.
\end{itemize}

\textbf{Federated learning server:} The FL server mainly coordinates the following processes: 
\begin{itemize}
    \item Client selection: in the default setting, clients are scheduled to participate update and aggregation process regularly. The backend will launch the user profiling and context quantization evaluation process for selected clients.
    \item Multi-client quantization planning: when all selected clients have completed the profiling and evaluation process, the server will filter the clients with precision levels with similar merits, and choose the optimal precision levels that maximize communication resource utilization in mixed-precision OTA aggregation.
    \item Mixed-precision OTA aggregation: The FL server aggregates model updates from clients of their current precision levels. After aggregation, the server will send the updated model back to clients along with the optimal precision levels for the next round for them to quantize the received model accordingly.
\end{itemize}

\subsection{RAG-based User Profiling}

The RAG-based user profiling process collects and infers user preferences on performance and operational contexts through a user-friendly conversational interface.

\begin{table*}[ht]
    \centering
    \caption{Examples of Contextual Factors and Inferable Factors}
    \label{tab:context_factors}
    \begin{tabular}{c|c|l} 
        \toprule
        \textbf{Contextual Factor} & \textbf{Inferable Factor} & \textbf{Examples} \\
        \midrule
        Device location & Input noise level & Bedroom $\rightarrow$ Low noise; Living room $\rightarrow$ High noise \\
        \midrule
        Interaction time & Input noise level, data quantity & Daytime $\rightarrow$ High noise, High quantity; Nighttime $\rightarrow$ Low noise, Low quantity \\
        \midrule
        Interaction frequency & Data quantity & High frequency $\rightarrow$ High quantity \\
        \midrule
        Task Type & Data distribution & Smart home hub $\rightarrow$ Short requests \\
        \bottomrule
    \end{tabular}
\end{table*}

\subsubsection{Contextual Factors}

The user preference for performance consists of three metrics: accuracy, energy consumption, and latency. These metrics are quantized by retrieving similar user cases from the knowledge database, based on the user's current feedback, operational contexts, and the estimated performance of their devices at the current precision levels. In comparison to conventional form-based feedback collection, RAG-LLM can analyse the user's sensitivity in these metrics through wording nuances in their feedback, prioritize primary user concerns in performance, and hence, facilitate accurate adjustments to meet their expectations. In addition, the RAG-LLM can analyse and link these sensitivities to operational contexts, as the same user could have different expectations and sensitivities in different scenarios.

Apart from supporting performance feedback, operational contexts are also indicators of the potential contribution of the client to the global model, see Table~\ref{tab:context_factors} for examples of such contextual factors and their potential effects. Factors such as data quality, quantity and distribution, which can be inferred from these contextual factors, are essential for the client contribution estimation.  FL service providers can use these inferred factors to estimate potential client contributions at different precision levels, and hence, select the optimal precision level for each client based on their learning strategies.

\subsubsection{RAG Database and LLM Integration} 

To support the LLM agents, we build a Context-Quantization-Feedback database, which stores the feedback from users of different contextual factors on performance at different quantization levels. When user feedback and contexts are collected via the chat interface, the LLM agent will retrieve similar user cases from the database, and estimate the user satisfaction, preference and client contribution at different precision levels based on the retrieved cases.

\subsubsection{User Profiling Pipeline}

The user profiling pipeline consists of the following steps:

\begin{enumerate}
    \item \textbf{Hardware specification extraction:} The backend extracts the hardware specification of the user device, including processor specs, RAM size, and power states.
    \item \textbf{Hardware quantization performance trade-off retrieval:} The backend queries the knowledge database for the quantization-performance trade-off on the same or similar hardware.
    \item \textbf{User interview feedback collection:} The agent prompts the user to provide feedback on the current performance and potential context changes since the last feedback collection, see Fig.~\ref{fig:framework-overview}-A for a chat example.
    \item \textbf{Contextual factor inference:} The LLM agent infers user preferences and contexts from past conversations.
    \item \textbf{User preference and contextual factor retrieval:} The agent retrieves similar user cases from the knowledge database with inferred user preferences and contexts.
    \item \textbf{User satisfaction and client contribution estimation:} The agent estimates the potential client contribution and user satisfaction at available precision levels based on retrieved contextual factors and hardware capability.
\end{enumerate}

\subsection{Context-Quantization Evaluation}

We define a reward-penalty model for determining the optimal quantization level for each client in a federated learning (FL) setting. The model considers multiple factors, each with an associated user-defined sensitivity weight. Assume:

\begin{itemize}
    \item $\mathcal{F}$: Set of factors (e.g., accuracy, energy cost, latency).
    \item $q$: Quantization level assigned to a client.
    \item $w_f$: Sensitivity weight of factor $f \in \mathcal{F}$, where $\sum_{f \in \mathcal{F}} w_f = 1$.
    \item $R_f(q)$: Reward obtained from operating at quantization level $q$ for factor $f$ (e.g., improved accuracy).
    \item $P_f(q)$: Penalty incurred by operating at quantization level $q$ for factor $f$ (e.g., energy consumption).
    \item $C_q$: Contribution multiplier for potential client contribution operating at quantization level $q$.
\end{itemize}

Then, the total reward and total penalty for quantization level $q$ are computed as the following weighted sums:

\begin{equation}\label{eq:R_total}
R_\text{Total}(q) = C_q \cdot \sum_{f \in \mathcal{F}} w_f \cdot R_f(q)
\end{equation}

\begin{equation}\label{eq:P_total}
P_\text{Total}(q) = \sum_{f \in \mathcal{F}} w_f \cdot P_f(q)
\end{equation}

\begin{equation}\label{eq:Satisfaction_Score}
\text{Satisfaction Score}(q) = R_\text{Total}(q) - P_\text{Total}(q)
\end{equation}

The optimization goal is to select the quantization level $q$ that maximizes the $\text{Satisfaction Score}$ defined as the difference between total reward and total penalty:

\begin{equation}\label{eq:q_star}
q^* = \arg \max_{q} \left( \text{Satisfaction Score}(q) \right)
\end{equation}


\section{Experiments}

\subsection{Experimental Setup}

We validate our proposed RAG-based precision planning framework on a federated smart voice assistant system with the Automatic Speech Recognition (AER) task. The federation consists of 100 simulated clients with diverse hardware capabilities and Gaussian distributed sensitivity to performance factors including accuracy, energy savings, and latency. We define the following experimental settings:

\textbf{Dataset and Model:}
The model structure is DeepSpeech2~\cite{amodei16deepspeech}, and the federated model is trained for 100 communication rounds. For client datasets, we filter the Common Voice dataset~\cite{commonvoice2020} with keywords related to the four main uses of smart voice assistants, \textit{Entertainment}, \textit{Smart Home}, \textit{General Query} and \textit{Personal Request}. We define these categories and their distribution, see Table~\ref{tab:data_distribution}, based on the usage statistics from the PWC research report~\cite{pwc_research}. 

\begin{table}[!t]
    \centering
    \caption{Smart Voice Assistant Data Distribution}
    \label{tab:data_distribution}
    \resizebox{\columnwidth}{!}{
    \begin{tabular}{c|c|c|c|c} 
        \toprule
        \textbf{Category} & \textbf{Entertainment} & \textbf{Smart Home} & \textbf{General Query} & \textbf{Personal Request} \\
        \midrule
        \textbf{Percentage} & 32.7\% & 16.0\% & 31.9\% & 19.4\% \\
        \bottomrule
    \end{tabular}
    }
\end{table}

\textbf{Metrics and Comparison:}
To showcase the advantage of our RAG-based user profiling precision planning framework, we compared it on the same federated learning system but with a unified standard precision planner, i.e., divide users in tiers by their hardware capabilities and assign the same precision level to each tier regardless of user preference and operational contexts. We measure the following metrics: 

\begin{itemize}
    \item \textbf{User Satisfaction Score}: the user satisfaction score defined in Equation~\ref{eq:Satisfaction_Score}. 
    \item \textbf{Relative Energy Cost}: we do not directly measure energy costs, instead, we measure the relative energy cost compared to the highest available precision level, and therefore the relative energy cost is a percentage below 100\%. 
    \item \textbf{Final Global Model Accuracy}: The final word accuracy of the global model after 100 communication rounds.  
\end{itemize}

\subsection{Results}

\subsubsection{User Satisfaction versus Energy Cost}

Our RAG-based user profiling precision planning framework generates personalized standards based on user preference and their operational contexts, resulting in a more accurate satisfaction estimate compared to the FL system that plans precision levels with unified standards, and the average satisfaction score is 10\% higher (0.66 compared to 0.60) while saving about 20\% energy. We also tested that when energy savings is the top priority of the mixed-precision FL system, our framework can trade 22\% average satisfaction score ((0.47 compared to 0.60)) for a total of 28\% energy savings.

\begin{figure}[htbp]
    \centering
    \includegraphics[width=0.9\linewidth]{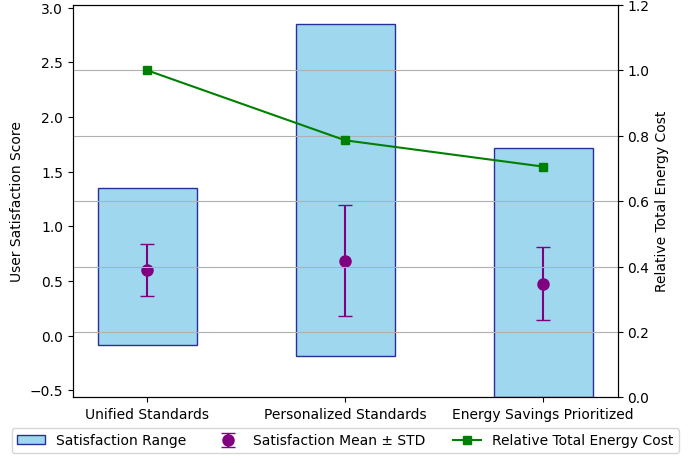}
    \caption{Distribution of User Satisfaction Scores and Relative Energy Cost. Compared to planning precision levels with unified standards, personalized standards can achieve 10\% higher average satisfaction score,  and 20\% of energy cost. When prioritise the federated system towards energy savings, 
    22\% satisfaction score can be traded for a total of 28\% energy saving.}
    \label{fig:satisfaction_energy_cost}
\end{figure}

\subsubsection{Global Model Performance}

\begin{figure}[htbp]
    \centering
    \includegraphics[width=0.9\linewidth]{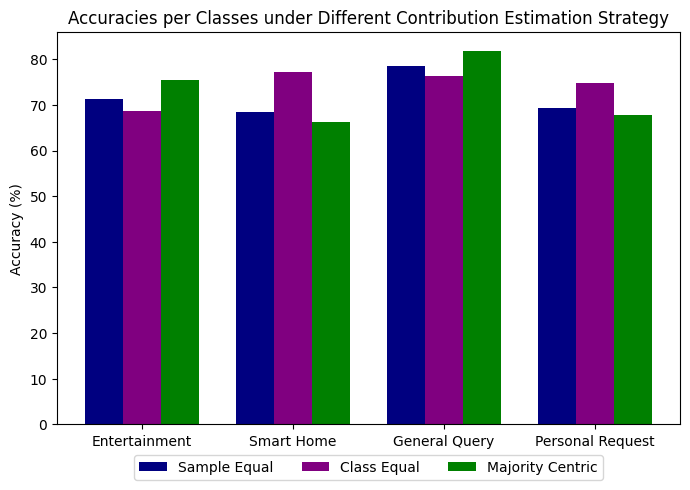}
    \caption{Word accuracy of the global model after 100 communication rounds by classes with different strategies. Compared to the default strategy, with b) class equal strategy, biased towards minority classes, our framework trades 2\% accuracy of the majorities for 5\% of that of the minorities; while with c) majority centric strategy, our framework extended the accuracies of majority classes by 4\% with 3\% lower accuracies for minority classes.}
    \label{fig:accuracy_contribution}
\end{figure}

Estimation of potential client contribution to the global model depends on the training strategy. We experimented with three different strategies with our framework: a) default FedAvg~\cite{McMahan2016FL} i.e. treat every sample equally; b) class equal strategy, attempts higher precision levels to samples of minority classes; c) majority centric strategy, attempts higher precision levels to samples of majority classes. Our RAG-based framework can estimate data distribution via contextual factors without breaching user privacy. Refer to actual data distribution in Table~\ref{tab:data_distribution}, see Fig.~\ref{fig:accuracy_contribution} compared to accuracies of FedAvg~\cite{McMahan2016FL}, our framework improved the accuracies of minority classes (smart home and personal request) and majority classes (entertainment and general query) with the corresponding biased strategies.

\section{Conclusion}

In this paper, we proposed a novel RAG-based user profiling for precision planning framework for Mixed-Precision Over-the-Air Federated Learning (MP-OTA-FL) systems, utilizing Retrieval-Augmented Generation (RAG)-powered Large Language Models (LLMs) for dynamic client profiling and quantization optimization. The proposed framework addresses key challenges in quantization-level selection and produces personalized precision planning through a conversational user profiling interface and dynamic RAG database utilization. Experimental evaluations demonstrated significant improvements in user satisfaction, energy savings, and global model accuracy compared to traditional quantization approaches with unified standards. Furthermore, our implementation is open-sourced to foster community-driven innovation in human-centred federated learning.

\bibliographystyle{IEEEtran}
\bibliography{references}

\end{document}